\documentclass[11pt,a4paper]{article}
\usepackage{emnlp2021}
\usepackage{times}
\usepackage{latexsym}

\usepackage{microtype}

\usepackage{url}
\usepackage[T5,T1]{fontenc}
\usepackage[utf8]{inputenc}
\usepackage{booktabs}
\usepackage{multirow}
\usepackage{multicol}
\usepackage{diagbox}
\usepackage{stfloats}

\usepackage{microtype}
\usepackage{graphicx}
\usepackage{subcaption}
\usepackage{booktabs} %
\usepackage{mathtools}
\usepackage{tikz}
\usetikzlibrary{bayesnet}
\usetikzlibrary{arrows}
\usetikzlibrary{backgrounds}
\usepackage{tikz-layers}

\usepackage{tabularx}
\usepackage{tabulary}
\usepackage{colortbl}
\usepackage{tabu}

\usepackage{amsmath}
\usepackage{amsfonts}
\usepackage{amssymb}
\usepackage{cuted}
\usepackage{dsfont}
\usepackage{soul}

\usepackage{algorithm}
\usepackage[noend]{algorithmic}
\usepackage{etoolbox}
\usepackage{xparse}

\usepackage[noabbrev,capitalise]{cleveref}
\newcolumntype{Y}{>{\centering\arraybackslash}X}

\usepackage{enumitem}
\newlist{todolist}{itemize}{2}
\setlist[todolist]{label=$\square$}
\usepackage{pifont}

\usepackage{todonotes}
\makeatletter
\newcommand*\iftodonotes{\if@todonotes@disabled\expandafter\@secondoftwo\else\expandafter\@firstoftwo\fi}  %
\makeatother

\definecolor{olive}{HTML}{3D9970}

\newcommand{\ignore}[2][]{}

\newcommand{\vtheta}{{\boldsymbol \vartheta}}

\newcommand{\xx}{\mathbf{x}}
\newcommand{\hh}{\mathbf{h}}

\DeclarePairedDelimiterX{\infdivx}[2]{(}{)}{%
  #1\;\delimsize\|\;#2%
}

\newcommand{\lv}[1]{{\boldsymbol #1}}

\definecolor{Gray}{gray}{0.92}
\definecolor{LightCyan}{rgb}{0.88,1,1}

\newcolumntype{a}{>{\columncolor{Gray}}c}

\title{Modelling Latent Translations for Cross-Lingual Transfer}

\author{Edoardo Maria Ponti$^{1, 2}$~~~~\textbf{Julia Kreutzer}$^{3}$~~~~\textbf{Ivan Vuli\'c}$^{4}$~~~~\textbf{Siva Reddy}$^{1, 2, 5}$ \\   
    $^1$Mila -- Quebec Artificial Intelligence Institute~~~$^2$McGill University\\
    $^3$Google Research~~~$^4$University of Cambridge~~~$^5$ Facebook CIFAR AI Chair \\
    $^1${\tt \{edoardo-maria.ponti,siva.reddy\}@mila.quebec} \\
    $^3${\tt jkreutzer@google.com}  ~~~$^4${\tt iv250@cam.ac.uk}\\
}

\date{}

\begin{document}
\maketitle
\begin{abstract}
While achieving state-of-the-art results in multiple tasks and languages, translation-based cross-lingual transfer is often overlooked in favour of massively multilingual pre-trained encoders. Arguably, this is due to its main limitations: \textbf{1)} translation errors percolating to the classification phase and \textbf{2)} the insufficient expressiveness of the maximum-likelihood translation. To remedy this, we propose a new technique that integrates both steps of the traditional pipeline (translation and classification) into a single model, by treating the intermediate translations as a latent random variable. As a result, \textbf{1)} the neural machine translation system can be fine-tuned with a variant of Minimum Risk Training where the reward is the accuracy of the downstream task classifier. Moreover, \textbf{2)} multiple samples can be drawn to approximate the expected loss across all possible translations during inference.  We evaluate our novel latent translation-based model on a series of multilingual NLU tasks, including commonsense reasoning, paraphrase identification, and natural language inference. We report gains for both zero-shot and few-shot learning setups, up to 2.7 accuracy points on average, which are even more prominent for low-resource languages (e.g., Haitian Creole). Finally, we carry out in-depth analyses comparing different underlying NMT models and assessing the impact of alternative translations on the downstream performance.

\end{abstract}

\section{Introduction}

Cross-lingual knowledge transfer supports the development of natural language technology for many of the world's languages \citep[\textit{inter alia}]{ruder2019survey,ponti-etal-2019-modeling}. The approach currently predominant for cross-lingual transfer relies on massively multilingual pre-trained encoders \citep{conneau-etal-2020-unsupervised,xue2020mt5,liu-etal-2020-multilingual} that are fine-tuned on a source language and perform zero-shot \citep{wu-dredze-2019-beto,10.1162/tacl_a_00374} or few-shot \citep{lauscher-etal-2020-zero,Zhao:2020arxiv} prediction in a target language.

An alternative approach for cross-lingual transfer, \textit{translate test}, is based on translating the evaluation set into the source language and leveraging a monolingual classifier instead \citep{banea-etal-2008-multilingual,durrett-etal-2012-syntactic,conneau-etal-2018-xnli}. This approach is currently under-investigated and usually relegated to the role of a baseline due to its lower flexibility, e.g., it is not suitable for sequence labelling tasks.
Yet, it achieves the state-of-the-art results in most benchmarks for multilingual Natural Language Understanding and Question Answering tasks \citep[\textit{inter alia}]{hu2020xtreme,ponti-etal-2020-xcopa,ruder2021xtreme}. Moreover, the availability of off-the-shelf translation models for multiple languages \citep{wu2016google,tiedemann-thottingal-2020-opus,liu-etal-2020-multilingual} provides coverage for transfer to a large number of target languages. Indeed, very recent preliminary results suggest that the translation-based transfer might even outperform \textit{monolingual} pre-trained models in languages different from English \citep{isbister2021stop}.

Translation-based transfer, however, currently suffers from two main limitations. First, the errors in translation accumulate along the pipeline. In fact, sentences that are possibly not faithful to the original in the target language and/or not grammatical in the source language are fed to the classifier, which degrades its performance. Second, only the maximum-likelihood translation is usually retrieved, which may not capture the precise meaning of the original sentence and its most relevant features for the downstream task.

In this work, we propose a method to amend these limitations and further enhance translation-based transfer. In particular, by treating the previously separate components for translation and classification as an integrated system, we re-interpret the traditional pipeline as a \textit{single model} with an intermediate \textit{latent translation} between the target text and its classification label. As a consequence of this change, 1) the machine translation component receives a feedback signal from the downstream loss and can be fine-tuned to better adapt to a specific task; 2) multiple translations can be sampled from the latent representation to perform ensemble prediction. Crucially, this method is sample efficient, as both components can be initialised with pre-trained models and deployed in a zero-shot or few-shot learning setting.

Na\"{\i}vely training the machine translation system via gradient descent, however, is often impossible due to the incompatibility of the token vocabularies of the two components. Therefore, we devise a universal method for fine-tuning that is suitable for any pair of pre-trained translator and classifier. We propose an optimisation scheme based on Minimum Risk Training \citep[MRT; ][\textit{inter alia}]{och-2003-minimum, smith-eisner-2006-minimum,shen-etal-2016-minimum}: it only requires the gradient of the translation scores and a reward based on downstream classification metrics.

Our evaluation is conducted on all multilingual Natural Language Understanding (NLU) tasks that are part of the popular XTREME \citep{hu2020xtreme} and XTREME-R \citep{ruder2021xtreme} cross-lingual transfer benchmarks. These include PAWS-X~\citep{yang-etal-2019-paws} for paraphrase identification, XCOPA \citep{ponti-etal-2020-xcopa} for commonsense reasoning, and XNLI \citep{conneau-etal-2018-xnli} for natural language inference. Our model improves over standard translation-based methods in zero-shot and few-shot scenarios, up to 2.6 accuracy points on average, with peaks of 5.6 points for resource-poor languages like Haitian Creole.

As an additional contribution, we also examine for the first time the impact of translation quality (as measured by BLEU) and multilingual coverage of several models on the downstream classification performance. In particular, we compare among the Google Cloud Translation API,\footnote{\url{https://cloud.google.com/translate}}
Marian MT ~\citep{tiedemann-thottingal-2020-opus,junczys-dowmunt-etal-2018-marian}, and mBART ~\citep{liu-etal-2020-multilingual,tang2020multilingual}, revealing substantial differences among these models. 
We release our code publicly at: \url{github.com/McGill-NLP/latent-translation}.

\section{Latent Translation Model}
\label{ss:model}
In the present work, we are concerned with the problem of performing zero/few-shot inference in any given target language $\ell_t$ by transferring knowledge from a source language $\ell_s$. Specifically, we focus on classification tasks with data of the form $\mathcal{D}_{\ell} \triangleq \{(\xx, y)\}_1^n$, where $\xx \in \Sigma_\ell^\star$ is a discrete sequence of tokens from the vocabulary $\Sigma_{\ell}$  and $y \in \mathbb{N}$ is a label index. We further assume that a parallel corpus $\mathcal{P} = \{(\xx_{\ell_s}, \xx_{\ell_t})\}_1^p$ is available.
This enables translation-based transfer \citep{banea-etal-2008-multilingual,durrett-etal-2012-syntactic}, which comes in two flavours: either the evaluation set can be translated into $\ell_s$ (\textit{translate test}), or the training set can be translated into $\ell_t$ (\textit{translate train}). We opt for the former, as it is both more efficient (due to $|\mathcal{D}_{eval}| \ll |\mathcal{D}_{train}|$) and effective \citep{conneau-etal-2018-xnli,hu2020xtreme}.

`Translate test' transfer relies on two main components: 1) a classifier $f_\vtheta$ parameterised by $\vtheta \in \mathbb{R}^d$ and trained on $\mathcal{D}_{\ell_s}$ and 2) a translator $f_\lv{\varphi}$ parameterised by $\lv{\varphi} \in \mathbb{R}^b$ trained on $\mathcal{P}$. These are deployed for predictive inference sequentially in the following pipeline: first, the $i$-th target language sentence(s) $\xx_{\ell_t}$ are mapped to their translation in the source language $\xx_{\ell_s}$. Afterwards, $\xx_{\ell_s}$ is fed to the classifier to produce the label $\hat{y}$.

However, this pipeline is arguably encumbered by two main limitations. Firstly, there is no information flow between the translator and the classifier; therefore, the errors in translation cannot be corrected in the subsequent step. Secondly, there may exist multiple correct translations, each reflecting different facets of the original sentence. Therefore, a single maximum-likelihood translation may not be representative of the underlying distribution, which is conceivably multi-modal.

\begin{figure}
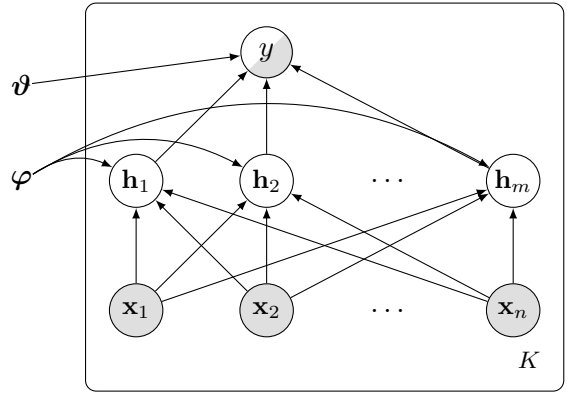

  \centering

  \tikz{ %
  \tikzset{>=latex}
  \tikzstyle{selector}=[-|, snake=snake,segment amplitude=.4mm,segment length=2mm] 

     \node[draw,circle] (k) {$y$}; %
  \node[latent,below=of k] (h2) {$\hh_2$};%
  \node[const,right=of h2] (dh) {$\dots$};%
\node[latent,right=of dh] (hn) {$\hh_m$};%
  \node[obs,below=of h2] (x2) {$\xx_2$}; %
  \node[latent,left=of h2] (h1) {$\hh_{1}$};%
  \node[obs,below=of h1] (x1) {$\xx_1$}; %
  \node[const,right=of x2] (dx) {$\dots$};%
  \node[obs,below=of hn] (xn) {$\xx_n$}; %
  \node[const,left=of h1] (phi) {$\lv\varphi$};%
  \node[const,above=of phi] (theta) {$\lv\vartheta$};%
  \begin{scope}[on behind layer]
      \fill[fill=gray!25] (k.225) arc [start angle=225, end angle=405, radius=3.5mm];
    \end{scope}

     \plate [inner sep=.3cm] {plate2} {(k) (x1) (hn)} {$K$}; %

     \edge {x1} {h1}
     \edge {x1} {h2}
     \edge {x1} {hn}
     \edge {x2} {h1}
     \edge {x2} {h2}
     \edge {x2} {hn}
     \edge {xn} {h1}
     \edge {xn} {h2}
     \edge {xn} {hn}
     \edge {h1} {k}
     \edge {h2} {k}
     \edge {hn} {k}
     \draw[->] (phi) to[bend left] (h1);
      \draw[->] (phi) to[bend left] (h2);
       \draw[->] (phi) to[bend left] (hn);
     \edge{theta}{k}
  }%
  \caption{A Bayesian graph of the generative model for latent translation cross-lingual transfer.}
  \label{fig:grfarch}
\end{figure}
Therefore, to grapple with both these problems, we propose to integrate both the translator and the classifier components into a unified model. This amounts to treating the translations as a latent random variable $\hh \in \Sigma_{\ell_s}^\star$ situated in between the target language sentences $\xx_{\ell_t}$ and the label $y$. From a Bayesian perspective, this is equivalent to the graphical model shown in \cref{fig:grfarch}. Hence, if we assume the conditional independence $(y \perp\!\!\!\perp \xx \mid \hh)$, posterior inference over the neural parameters $\vtheta$ and $\lv\varphi$ given $\mathcal{D}_{\ell_s}$ requires estimating:
\begin{align} \label{eq:posterior}
    p(\vtheta, \lv\varphi &\mid y, \xx)  \\ &\propto p(y \mid \xx, \vtheta, \lv\varphi) \; p(\vtheta) \; p(\lv\varphi)  \nonumber \\
    &= \sum_{\hh \in \Sigma^\star_\ell} p(y \mid \hh, \vtheta) \; p(\hh \mid \xx, \lv\varphi) \; p(\vtheta) \; p(\lv\varphi). \nonumber
\end{align}
In other words, the latent variable $\hh$ must be integrated out. By virtue of \cref{eq:posterior}, the estimate for the translator parameters $\lv\varphi$ is influenced by the label $y$. Hence, out-of-the-box translation models can be adapted to the domain- or task-specific cues based on the feedback that downstream classification provides for any translation that they generate. Moreover, the entirety of the space $\Sigma^\star_\ell$ is explored rather than just the maximum-likelihood sequence. Thus, the multi-faceted semantics of the input sentence in the target language is better preserved.

This formulation, however, poses additional challenges. First, the domain $\Sigma^\star_\ell$ is countably infinite. Therefore, integrating over this space, and estimating the full probability distribution, is virtually impossible. Second, being $\hh$ discrete sequences, the model is not fully differentiable. Hence, training it via gradient descent is not trivial. In what follows, we propose approximate solutions in order to perform inference under our model.

\subsection{Monte Carlo Sampling of Translations}
\label{ssec:mcmc}
While it may not be feasible to integrate over the space of all possible translations, we can approximate the likelihood term of \cref{eq:posterior} through a finite set of $k$ Monte Carlo samples:
\begin{align} \label{eq:likelihood}
    \sum_{\hh \in \Sigma^\star_\ell} &p(y \mid \hh, \vtheta) \; p(\hh \mid \xx, \lv\varphi) \\ 
    &= \mathop{\mathbb{E}}_{\hh \sim p(\cdot \mid \xx, \lv\varphi)} p(y \mid \hh, \vtheta) \nonumber \\
    &\approx \frac{1}{k} \sum_{i=1}^k p(y \mid \hh_i, \vtheta) \quad \hh_i \sim p(\cdot \mid \xx, \lv\varphi) \nonumber
\end{align}
In practice, this amounts to performing an ensemble prediction,\footnote{In addition to Monte Carlo sampling, we also experimented with weighted averages, where the weight was the normalised sample probability. However, we found this to be detrimental to downstream performance.} where $k$ candidate outputs of the translator $\hh_i \sim p(\cdot \mid \xx, \lv\varphi)$ are fed to the classifier. The predictive distributions yielded by the classifier given each candidate are then averaged:
\begin{align} \label{eq:clsloss}
    \mathcal{L}(\vtheta) &= \min_{\vtheta} \sum_{i=1}^n - \log \left[ \frac{1}{k} \sum_{j=1}^k \sigma\bigl(f_\vtheta(\hh_i)\bigr) \right]_{y_i} \nonumber \\ &\textrm{where} \quad \hh_j \sim f_\lv{\varphi}(\xx_i)
\end{align}
and $y_i$ indexes to the probability of the gold label of the $i$-th example and $\sigma$ is the softmax function.

\subsection{Minimum Risk Training}
\label{ssec:mrt}
The second difficulty of our integrated model lies in the fact that the latent translations are discrete sequences, which implies a non-differentiable hard decision boundary. This could be easily addressed by relaxing the tokens that are part of the translation into continuous variables \citep{maddison2017concrete,jang2017categorical} or adopting straight-through estimators \citep{bengio2013estimating,raiko2014techniques}. Nonetheless, this may still be impractical. In fact, it is common to instantiate both the translator $f_\lv{\varphi}$ and the classifier $f_\vtheta$ with pre-trained neural models. In this case, the output vocabulary of the former often does not coincide with the latter, due to different sub-word tokenisation strategies \citep{rust2020good}. Therefore, the domain $\Sigma^\star_{\ell_s}$ of the translator output and the classifier input may also not correspond.

In order to make our method as general as possible and match any pair of translator and classifier with arbitrary vocabularies, we resort instead to a reinforcement learning technique to fine-tune the translator parameters. In particular, we adopt a version of Minimum Risk Training \citep[MRT;][]{shen-etal-2016-minimum}: its key idea is minimising the risk (expressed as a negative reward weighted by its probability). MRT is typically harnessed for NMT as a downstream task; the reward is thus BLEU or similar metrics. However, in our setting we propose to use classification accuracy as a reward. Let $p^\star(\cdot)$ represent the score (i.e.\ the unnormalised probability) of a translation. The loss can be formulated as follows:
\begin{align} \label{eq:mrt}
   \mathcal{L}(\lv\varphi) = \min_\lv\varphi \sum_{i=1}^n \sum_{j=1}^k \frac{ p^\star(\hh_{ij} \mid \lv\varphi, \xx_i)}{\sum_{j^\prime} p^\star(\hh_{ij^\prime} \mid \lv\varphi, \xx_i)} \\ \times \, - \log p(y_i \mid {\hh_{ij}}, \vtheta), \nonumber
\end{align}
where $n$ is the number of training inputs (here: few-shot learning), $k$ the number of translation samples generated for each input, $\hh$ a latent translation, and $\lv\varphi$ the MT model parameters. The downstream task reward is $\mathcal{R}(\hh_{ij}) \triangleq \log p(y_i \mid {\hh_{ij}}, \vtheta)$, the log-likelihood of the classifier prediction based on the $j$-th individual translation.\footnote{Larger sample sizes $k$ encourage the MT model to explore more alternative translations and yield closer approximations to the true expected risk, but are expensive to compute.
}

We optimise the parameters $\lv\varphi$ through gradient descent, where the gradient of the loss in \cref{eq:mrt} with respect to the $z$-th weight $\lv\varphi_z$ is computed as:
\begin{align}
    \frac{\partial \, \mathcal{L}(\lv\varphi)}{\partial \, \lv\varphi_z} = & \sum_{i=1}^n \mathop{\mathbb{E}}_{\hh_{i} \mid \lv\varphi, \xx_i}     \biggl[\frac{\partial \, p^\star(\hh_{i} \mid \lv\varphi, \xx_i) / \partial \, \lv\varphi_z}{p^\star(\hh_{i} \mid \lv\varphi, \xx_i)} \nonumber \\ \times  &\biggl(\mathcal{R}(\hh_{i}) - 
    \mathop{\mathbb{E}}_{\hh_{i^\prime} \mid \lv\varphi, \xx_i} 
    \mathcal{R}(\hh_{i^\prime}) \biggr)\biggr]
\end{align}
where the expectations are computed by explicitly enumerating the $k$ Monte Carlo samples. 
\subsection{MAP Inference}
\label{ss:inference}

Combining the objectives outlined in \cref{ssec:mcmc} and \cref{ssec:mrt}, we finally obtain the maximum-a-posteriori (MAP) approximation to perform posterior inference over the graphical model in \cref{fig:grfarch}. This is expressed in the following objective, which we use for fine-tuning the classifier and translator parameters during few-shot learning:
\begin{equation}
   \min_{\vtheta, \lv\varphi} \mathcal{L}(\vtheta) + \mathcal{L}(\lv\varphi) + \frac{\lambda}{2}||\vtheta||_2^2 + \frac{\lambda}{2} ||\lv\varphi||_2^2,
   \label{eq:map}
\end{equation}
where $\mathcal{L}(\vtheta)$ is taken from \cref{eq:clsloss} and $\mathcal{L}(\lv\varphi)$ is taken from \cref{eq:mrt}. Note that \cref{eq:map} contains two regularisers, which correspond to the prior terms in \cref{eq:posterior}. In particular, we take $\vtheta \sim \mathcal{N}({\bf 0}, I\lambda^{-1})$ and $\lv\varphi \sim \mathcal{N}({\bf 0}, I\lambda^{-1})$.

In zero-shot setups, and after that the parameters have been fine-tuned in few-shot setups, we perform predictive inference on new data points in the evaluation set through the ensemble of Monte Carlo samples, as described in \cref{eq:likelihood}.

\section{Experimental Setup}
\label{sec:exp}
\noindent \textbf{Evaluation Tasks and Data.} 
We conduct experiments on three established cross-lingual transfer datasets for natural language understandings tasks. 1) PAWS-X~\citep{yang-etal-2019-paws} for paraphrase identification: given a pair of sentences, a binary label specifies whether they express an identical meaning; 2) XCOPA \citep{ponti-etal-2020-xcopa} for commonsense causal reasoning: given a premise, a question, and a pair of (cause, effect) hypotheses, the model must determine which of the two is correct; 3) XNLI \citep{conneau-etal-2018-xnli} for natural language inference: a pair of sentences is classified as either an entailment, a contradiction, or a neutral relationship. Together, these 3 tasks cover a wide variety of typologically diverse languages (22 distinct ones in addition to English). 

Following prior work \cite{hu2020xtreme,ruder2021xtreme}, English is the source language in all experiments. In all three tasks, the English training set is used to train the classifier, the English development set for hyper-parameter selection, the development sets in other languages for few-shot learning (i.e., for fine-tuning both the classifier and translator), and the test set of the target languages for evaluation. 

\vspace{1.5mm}
\noindent \textbf{Machine Translation Systems.}
In order to assess the impact of the underlying translation model on downstream performance, we compare three established NMT system: 1) a closed-source software, the Google Cloud Translation API.\footnote{As of 6 October 2020 for XCOPA and 30 April 2021 for PAWS-X and XNLI.} Moreover, we consider two open-source systems: 2) mBART~\citep{liu-etal-2020-multilingual,tang2020multilingual}, a multilingual model covering 50 languages pre-trained on a denoising objective and fine-tuned on parallel data. 3) 
Marian MT~\citep{junczys-dowmunt-etal-2018-marian}, a set of hundreds of pair-wise models which were directly trained on parallel data from OPUS~\citep{tiedemann-thottingal-2020-opus}.\footnote{Pre-trained models are sourced from \url{github.com/huggingface/transformers}: for Marian \texttt{Helsinki -NLP/opus-mt-\{src\}-\{tgt\}} whereas for mBART \texttt{facebook/mbart-large-50-many-to-one-mmt}.} 

mBART is an encoder-decoder model where both the encoder and decoder have $12$ Transfomer layers and $16$ attention heads per layer. The hidden dimension is $1024$, whereas the FFN inner dimension is $4096$. Marian MT is a lighter model, where all the parameters are exactly halved. Further, Marian MT varies from mBART slightly in other regards: it employs static (sinusoid) positional embeddings and does not perform layer normalisation.

Several languages in our set of cross-lingual tasks are covered by some translation systems: no system currently covers \textsc{qu};\footnote{We refer to languages through their ISO 639-1 code.} mBART also lacks \textsc{bg}, \textsc{el}, and \textsc{ht}, Marian MT lacks \textsc{el}, \textsc{sw}, and \textsc{ta}.\footnote{ \citet{artetxe-etal-2020-translation} noted that the joint translation of all sentences that are part of a same example (e.g., premise and hypothesis) might be beneficial, as this retains their lexical overlap.
Despite this, in our implementation,
they are translated separately, as Marian and mBART do not handle multiple inputs simultaneously.
}

\vspace{1.5mm}
\noindent \textbf{Classifier.}
As a classifier $f_\vtheta$ for the output of all MT systems, we use RoBERTa Large \citep{liu2019roberta}, a $24$-layer monolingual pre-trained encoder for English, with a 2-layer perceptron (MLP) head. The encoder's hidden size is $1024$, whereas the internal representation of the MLPs (both in the encoder and the head) is $1024$.

In order to establish another common (and not translation-based) baseline in all evaluation tasks, we also fine-tune a \textit{multilingual} encoder with an identical configuration to RoBERTa, XLM-R Large \citep{conneau-etal-2020-unsupervised}.
In this case, the target language text is fed directly to the classifier, without requiring translation. We label this approach \textit{ME}, as opposed to  `translate test' (\textit{TT}).\footnote{We also experimented with using massively multilingual NMT models, such as mBART, as encoders for ME-style transfer \citep{eriguchi2018zero,siddhant2020evaluating}. However, their scores significantly lag behind XLM-R Large. For brevity, we report them in \cref{tab:extra_res} in Appendix.}

\begin{figure}[t]
    \centering
    \includegraphics[width=0.97\columnwidth]{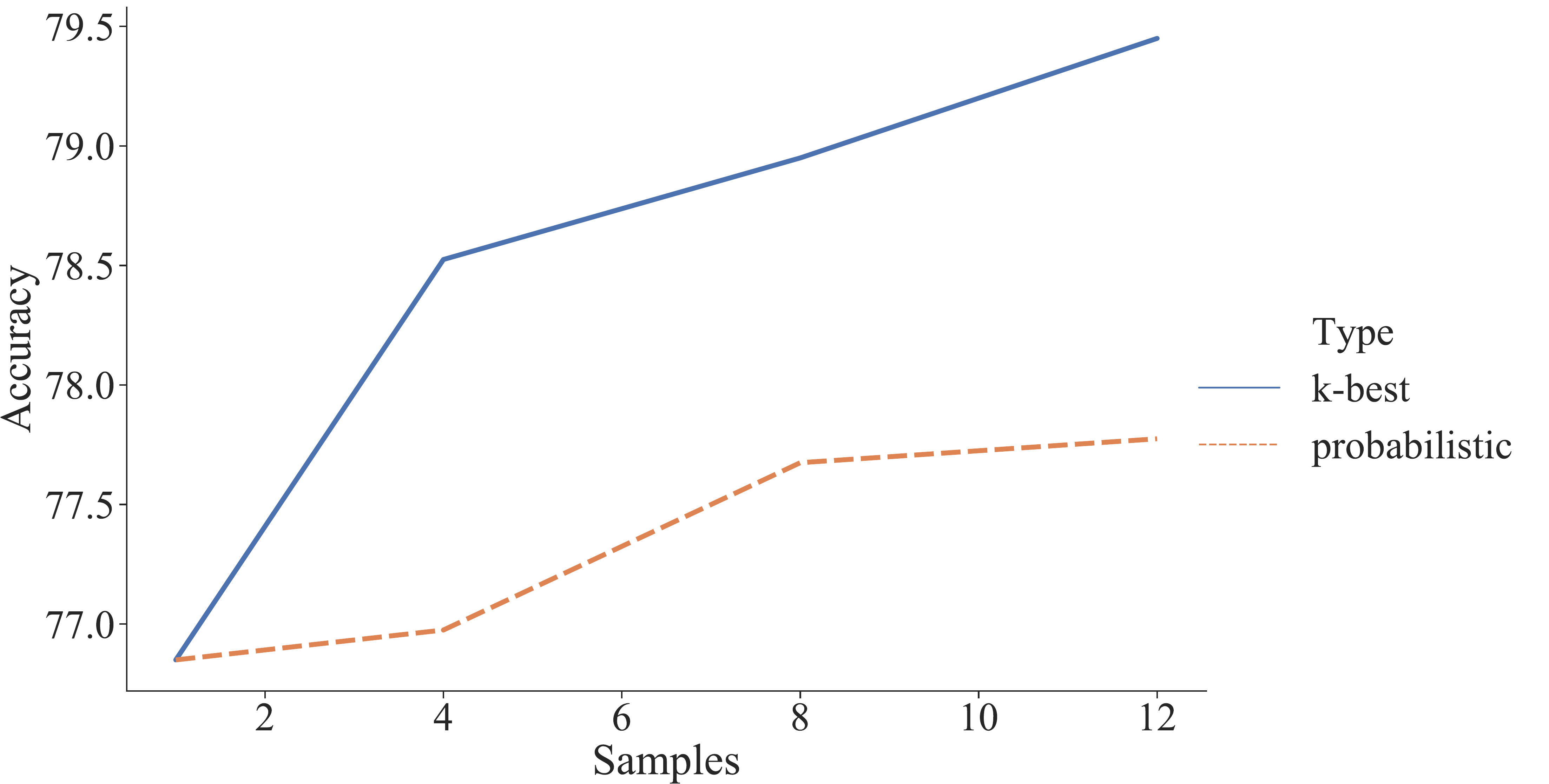}
    \vspace{-1.5mm}
    \caption{Average cross-lingual performance of ensemble prediction based on different numbers of samples.}
    \label{fig:sampling}
    \vspace{-1.5mm}
\end{figure}

\vspace{1.5mm}
\noindent \textbf{Optimisation.}
Both during fine-tuning on English and few-shot learning on the target language, we train all models for $5$ epochs. The classifier's parameters are optimised through Adam \citep{kingma2014adam} with learning rate $8 \times 10^{-6}$ and $\epsilon = 10^{-8}$, whereas the translator's parameters through SGD with learning rate $10^{-2}$. We use a dropout of $0.1$ during fine-tuning, and clip the gradient norm to $1$. The maximum sequence length is $128$, and the batch size is $24$. Finally, for translation sampling we select the $k$ most likely sequences with a beam size of $12$ and a temperature of $1$. We verified empirically that probabilistic sampling performs worse (cf.\ \cref{fig:sampling}).

\begin{table*}[p]

\centering
\subcaptionbox{XCOPA: zero-shot learning.\label{tab:xcopa-zero}}{
{\footnotesize
\begin{tabularx}{\columnwidth}{>{\sc}Y|a|YYY|Y}
\toprule
& \multicolumn{1}{a|}{\textit{ME}} & \multicolumn{4}{c}{\textit{TT}: RoBERTa +} \\
\hline
    & \rotatebox[origin=Y]{270}{\bf XLM-R} & \rotatebox[origin=Y]{270}{\bf mBART} & \rotatebox[origin=Y]{270}{\bf Google} & \rotatebox[origin=Y]{270}{\bf Marian} & \rotatebox[origin=Y]{270}{\bf Marian} \\
& \rotatebox[origin=Y]{270}{L} & \rotatebox[origin=Y]{270}{1} & \rotatebox[origin=Y]{270}{1} & \rotatebox[origin=Y]{270}{1} & \rotatebox[origin=Y]{270}{12} \\
    \hline
en  & 85.4  &   89.8    & 89.8   & 89.8   & 89.8                         \\
\hline
et  & 72.6  & 69.8  & 82.2   & 83.4   & 84.4                         \\
ht  & $^\star$     & $^\star$     & 75.4   & 56.0     & 61.6                         \\
id  & 81.8  & 80.6  & 83.8   & 82.2   & 85.2                         \\
it  & 77.0    & 74.0    & 85.8   & 80.6   & 79.8                         \\
qu  & $^\star$     & $^\star$     & $^\star$      & $^\star$      & $^\star$                            \\
sw  & 62.8  & 50.2  & 76.6   & $^\star$      & $^\star$                            \\
ta  & 71.2  & 69.8  & 81.8   & $^\star$      & $^\star$                            \\
th  & 72.4  & 62.6  & 76.4   & 74.6   & 77.2                         \\
tr  & 71.6  & 74.6  & 83.4   & 79.6   & 83.0                           \\
vi  & 77.6  & 76.2  & 83.0   & 76.0     & 79.2                         \\
zh  & 80.2  & 82.8  & 85.2   & 82.4   & 85.2                         \\
\hline
\textbf{\normalfont{avg}} & 74.1  & 71.2  & 81.4   & 76.9   & 79.5               \\
\bottomrule
\end{tabularx}
}
}%
\subcaptionbox{XCOPA: few-shot learning.\label{tab:xcopa-few}}{
{\footnotesize
\begin{tabularx}{\columnwidth}{>{\sc}Y|a|YYY|YY}
\toprule
& \multicolumn{1}{a|}{\textit{ME}} & \multicolumn{5}{c}{\textit{TT}: RoBERTa +} \\
\hline
    & \rotatebox[origin=Y]{270}{\bf XLM-R} & \rotatebox[origin=Y]{270}{\bf mBART} & \rotatebox[origin=Y]{270}{\bf Google} & \rotatebox[origin=Y]{270}{\bf Marian} & \rotatebox[origin=Y]{270}{\bf Marian}  & \rotatebox[origin=Y]{270}{\bf + MRT} \\
& \rotatebox[origin=Y]{270}{L} & \rotatebox[origin=Y]{270}{1} & \rotatebox[origin=Y]{270}{1} & \rotatebox[origin=Y]{270}{1} & \rotatebox[origin=Y]{270}{12}  & \rotatebox[origin=Y]{270}{12} \\
\hline
en  & 87.4  & 89.6  & 89.6   & 89.6   & 89.6                         & 89.6                  \\
\hline
et  & 73.8  & 69.8  & 82.0     & 85.4   & 86.4                         & 86.6                   \\
ht  & $^\star$     & $^\star$     & 79.0     & 56.2   & 61.4                         & 61.0                    \\
id  & 83.0    & 80.8  & 84.4   & 86.2   & 87.2                         & 87.4                  \\
it  & 77.0    & 78.6  & 86.2   & 85.2   & 84.2                         & 86.2                    \\
qu  & $^\star$     & $^\star$     & $^\star$      & $^\star$      & $^\star$                            & $^\star$                     \\
sw  & 60.8  & 47.8  & 77.4   & $^\star$      & $^\star$                            & $^\star$                     \\
ta  & 72.8  & 72.2  & 80.0     & $^\star$      & $^\star$                            & $^\star$                     \\
th  & 75.0    & 62.0    & 82.6   & 76.2   & 78.4                         & 79.8                  \\
tr  & 73.0    & 75.4  & 82.4   & 81.4   & 83.0                           & 83.8                  \\
vi  & 76.2  & 77.2  & 82.0     & 79.0     & 78.0                           & 81.2                    \\
zh  & 80.8  & 81.6  & 85.6   & 84.8   & 84.0                           & 85.2                  \\
\hline
\textbf{\normalfont{avg}} & 74.7  & 71.7  & 82.2   & 79.3   & 80.3                         & 81.4       \\
\bottomrule
\end{tabularx}
}
}%
\vspace{3mm}

\centering
\subcaptionbox{PAWS-X: zero-shot learning.\label{tab:pawsx-zero}}{
{\footnotesize
\begin{tabularx}{\columnwidth}{>{\sc}Y|a|YYY|Y}
\toprule
& \multicolumn{1}{a|}{\textit{ME}} & \multicolumn{4}{c}{\textit{TT}: RoBERTa +} \\
\hline
    & \rotatebox[origin=Y]{270}{\bf XLM-R} & \rotatebox[origin=Y]{270}{\bf mBART} & \rotatebox[origin=Y]{270}{\bf Google} & \rotatebox[origin=Y]{270}{\bf Marian} & \rotatebox[origin=Y]{270}{\bf Marian} \\
& \rotatebox[origin=Y]{270}{L} & \rotatebox[origin=Y]{270}{1} & \rotatebox[origin=Y]{270}{1} & \rotatebox[origin=Y]{270}{1} & \rotatebox[origin=Y]{270}{12} \\
\midrule
en  & 95.75 & 95.99 & 95.99  & 95.99  & 95.99                        \\
\hline
de  & 90.60  & 89.54 & 91.25  & 91.05  & 91.40                         \\
es  & 91.60  & 87.79 & 92.05  & 91.45  & 91.80                         \\
fr  & 92.30  & 89.94 & 92.20   & 91.40   & 91.90                         \\
ja  & 81.59 & 77.49 & 81.09  & 72.89  & 74.54                        \\
ko  & 83.04 & 74.59 & 81.49  & 73.04  & 73.24                        \\
zh  & 84.34 & 82.04 & 85.24  & 82.44  & 82.64                        \\
\hline
\textbf{\normalfont{avg}} & 87.24 & 83.57 & 87.22  & 83.71  & 84.25  \\
\bottomrule
\end{tabularx}
}
}%
\subcaptionbox{PAWS-X: few-shot learning.\label{tab:pawsx-few}}{
{\footnotesize
\begin{tabularx}{\columnwidth}{>{\sc}Y|a|YYY|YY}
\toprule
& \multicolumn{1}{a|}{\textit{ME}} & \multicolumn{5}{c}{\textit{TT}: RoBERTa +} \\
\hline
    & \rotatebox[origin=Y]{270}{\bf XLM-R} & \rotatebox[origin=Y]{270}{\bf mBART} & \rotatebox[origin=Y]{270}{\bf Google} & \rotatebox[origin=Y]{270}{\bf Marian} & \rotatebox[origin=Y]{270}{\bf Marian}  & \rotatebox[origin=Y]{270}{\bf + MRT} \\
& \rotatebox[origin=Y]{270}{L} & \rotatebox[origin=Y]{270}{1} & \rotatebox[origin=Y]{270}{1} & \rotatebox[origin=Y]{270}{1} & \rotatebox[origin=Y]{270}{12}  & \rotatebox[origin=Y]{270}{12} \\
\midrule
en  & 97.10  & 96.85 & 96.85  & 96.85  & 96.85                        & 96.85                 \\
\hline
de  & 92.85 & 91.95 & 93.55       & 93.05  & 92.55                        &   93.40                    \\
es  & 93.20  & 91.45 & 93.25       & 92.80   & 94.10                         & 93.60                      \\
fr  & 93.35 & 92.20  &  93.55      & 93.55  & 93.70                         &  93.55                     \\
ja  & 85.19 & 83.04 & 82.94       & 79.54  & 81.19                        &  81.39                     \\
ko  & 85.69 & 79.39 & 85.54       & 80.34  & 80.49                        & 80.54                      \\
zh  & 87.19 & 85.34 &  88.44      & 86.44  & 87.04                        &  87.49                     \\
\hline
\normalfont{avg}  & 89.58 &	87.23	& 89.55 &	87.62 &	88.18 & 88.33
  \\
\bottomrule
\end{tabularx}
}
}%
\vspace{3mm}

\centering
\subcaptionbox{XNLI: zero-shot learning.\label{tab:xnli-zero}}{
{\footnotesize
\begin{tabularx}{\columnwidth}{>{\sc}Y|a|YYY|Y}
\toprule
& \multicolumn{1}{a|}{\textit{ME}} & \multicolumn{4}{c}{\textit{TT}: RoBERTa +} \\
\hline
    & \rotatebox[origin=Y]{270}{\bf XLM-R} & \rotatebox[origin=Y]{270}{\bf mBART} & \rotatebox[origin=Y]{270}{\bf Google} & \rotatebox[origin=Y]{270}{\bf Marian} & \rotatebox[origin=Y]{270}{\bf Marian} \\
& \rotatebox[origin=Y]{270}{L} & \rotatebox[origin=Y]{270}{1} & \rotatebox[origin=Y]{270}{1} & \rotatebox[origin=Y]{270}{1} & \rotatebox[origin=Y]{270}{12} \\
\midrule
en  & 88.84 & 91.24 & 91.24  & 91.24  & 91.24                        \\
\hline
ar  & 79.58 & 72.83 & 82.27  & 78.60   & 79.98                        \\
bg  & 83.21 & $^\star$     & 85.21  & 84.43  & 85.01                        \\
de  & 82.97 & 82.71 & 85.45  & 84.49  & 85.37                        \\
el  & 82.03 & $^\star$     & 84.09  & $^\star$      & $^\star$                            \\
es  & 84.27 & 78.56 & 86.88  & 85.73  & 86.44                        \\
fr  & 82.95 & 82.35 & 85.33  & 84.51  & 85.09                        \\
hi  & 76.48 & 73.25 & 77.26  & 63.71  & 65.14                        \\
ru  & 79.34 & 77.98 & 82.23  & 79.68  & 81.13                        \\
sw  & 72.19 & 34.66 & 75.12  & $^\star$      & $^\star$                            \\
th  & 76.92 & 45.28 & 77.40   & 74.07  & 75.34                        \\
tr  & 78.94 & 74.15 & 81.57  & 79.88  & 80.57                        \\
ur  & 72.57 & 60.55 & 71.79  & 55.3.0   & 55.44                        \\
vi  & 80.12 & 75.86 & 81.79  & 77.70   & 78.80                         \\
zh  & 80.00    & 78.34 & 81.73  & 79.02  & 79.78                        \\
\hline
\normalfont{avg} & 80.03 & 71.37 & 81.96  & 78.33  & 79.18           \\   
\bottomrule
\end{tabularx}
}
}%
\subcaptionbox{XNLI: few-shot learning.\label{tab:xnli-few}}{
{\footnotesize
\begin{tabularx}{\columnwidth}{>{\sc}Y|a|YYY|YY}
\toprule
& \multicolumn{1}{a|}{\textit{ME}} & \multicolumn{5}{c}{\textit{TT}: RoBERTa +} \\
\hline
    & \rotatebox[origin=Y]{270}{\bf XLM-R} & \rotatebox[origin=Y]{270}{\bf mBART} & \rotatebox[origin=Y]{270}{\bf Google} & \rotatebox[origin=Y]{270}{\bf Marian} & \rotatebox[origin=Y]{270}{\bf Marian}  & \rotatebox[origin=Y]{270}{\bf + MRT} \\
& \rotatebox[origin=Y]{270}{L} & \rotatebox[origin=Y]{270}{1} & \rotatebox[origin=Y]{270}{1} & \rotatebox[origin=Y]{270}{1} & \rotatebox[origin=Y]{270}{12}  & \rotatebox[origin=Y]{270}{12} \\
\midrule
en & 89.54 & 91.24 & 91.24 & 91.24 & 91.24 & 91.24 \\
\hline
ar & 81.83 & 79.78 & 84.73 & 82.61 & 83.41 & 83.57
 \\
bg & 85.49 & $^\star$ & 87.36 & 86.46 & 87.20 & 87.2
 \\
de & 84.59 & 85.55 & 87.20 & 86.11 & 86.86 & 87.5
 \\
el & 83.95 & $^\star$ & 86.42 & $^\star$ & $^\star$ & $^\star$ \\
es & 85.77 & 81.41 & 87.66 & 87.68 & 88.08 & 88.26
 \\
fr & 84.87 & 84.97 & 86.84 & 86.42 & 86.62 & 86.76
 \\
hi & 79.64 & 78.84 & 80.67 & 72.73 & 74.67 & 74.97
 \\
ru & 82.85 & 82.15 & 84.52 & 83.43 & 84.15 & 84.51
 \\
sw & 76.12 & 41.63 & 79.70 & $^\star$ & $^\star$ & $^\star$ \\
th & 80.52 & 53.30 & 81.07 & 79.84 & 80.83 & 80.95
 \\
tr & 81.35 & 79.50 & 84.28 & 83.75 & 84.31 & 84.33
 \\
ur & 76.30 & 69.93 & 77.14 & 63.63 & 64.22 &
65.56
 \\
vi & 83.03 & 81.07 & 84.11 & 81.85 & 83.47 & 82.97
 \\
zh & 82.35 & 82.45 & 84.25 & 82.67 & 84.19 & 84.13
 \\
\hline
\normalfont{avg} & 82.05 & 75.05 & 84.00 & 81.43 & 82.34 & 82.55 \\
\bottomrule
\end{tabularx}
}
}%
\caption{Results (Accuracy$\times$100) in zero-shot (left columns) and few-shot (right columns) scenarios for XCOPA (top tables), PAWS-X (centre tables), and XNLI (bottom tables). {$^\star$}MT system does not cover the target language. The numbers after each MT system refer to the number of translation samples $k$ (see \cref{sec:exp}).}
\label{tab:allres}
\end{table*}

\section{Results and Discussion}
The main results on XCOPA, PAWS-X, and XNLI, both in zero-shot and few-shot transfer scenarios, are summarised in Table~\ref{tab:allres}. 
In addition to the baselines, we report results on Monte Carlo sampling and MRT only for the best performing open-source model, Marian.\footnote{Moreover, these results are not available for Google MT because the API does not allow for multiple translations or fine-tuning.}
The scores offer multiple axes in comparison, discussed in what follows.%

\vspace{1.5mm}
\noindent \textbf{Multilingual Encoders versus \textit{Translate Test}.} Inspecting the global trends in Table~\ref{tab:allres}, the results mostly corroborate the received wisdom from prior work \cite{hu2020xtreme,ruder2021xtreme,ponti-etal-2020-xcopa}: \textit{translate test} coupling Google MT with a monolingual English encoder yields stronger cross-lingual transfer performance on average than using a state-of-the-art massively multilingual encoder such as XLM-R Large. However, while the finding is true when using Google MT, we note that 1) it does not hold across the board with other NMT systems, and 2) ME-based transfer  with XLM-R Large is confirmed as a strong non-MT transfer baseline. %
For instance, 1-best mBART falls behind XLM-R Large in all three tasks, and likewise for 1-best Marian MT in PAWS-X and XNLI. %

\vspace{1.5mm}
\noindent \textbf{A Comparison of NMT Models (1-best).}
Task performance varies dramatically according to the chosen MT system; what is more, it can serve as a (non-ideal) proxy of MT system quality. Google translations are by far the best (compare the average scores at the bottom of Table~\ref{tab:allres}), with pronounced gains over the two competitors in all three tasks, especially in zero-shot setups. Marian MT also displays significantly better results than mBART across the board, especially on XCOPA and XNLI, despite its smaller parameter count. Arguably, this is caused by the fact that Marian MT has separate models available for each language pair, whereas mBART is massively multilingual. This effect is known as the `curse of multilinguality' \citep{conneau-etal-2020-unsupervised}. Finally, while Marian MT reduces the gap to Google in few-shot transfer, Google remains the strongest alternative in this setup, too, for all three tasks.

\vspace{1.5mm}
\noindent \textbf{Multiple Samples and MRT.}
Our latent-based translation approach yields consistent gains over the base 1-best MT system in all three tasks: the improvements of 12-best Marian over its 1-best variant are observed in all zero-shot and few-shot runs. The further inclusion of MRT in few-shot setups results in additional small but consistent boosts, again in all evaluation tasks. This confirms that both components, as discussed in \cref{ss:model}, indeed focus on mitigating distinct limitations of the standard translation-based approach, and thus offer complementary benefits to the final task performance. Using multiple samples with MRT, an initially weaker NMT system can recover several points of performance, even outperforming an initially stronger NMT system for some tasks and languages. For instance, MRT with 12 samples is the peak-scoring variant for \textsc{tr} by 2 points, despite ``starting'' 2 points behind the Google 1-best baseline. We observe similar trends, among others, for \textsc{es} and \textsc{fr} on PAWS-X, or for \textsc{ru} and \textsc{zh} on XNLI. %
In sum, this implies that latent translation transfer can enhance existing models even with zero or few examples available, at the expense of a modest increase in inference time (cf.\ \cref{ss:inference}).\footnote{In addition to MRT, we considered fine-tuning the translator through self-training \citep{sennrich-etal-2016-improving} and learning a re-ranker for weighted ensemble prediction \citep{dong-etal-2017-learning-paraphrase}. Compared to MRT, both yield sub-par results, which are reported in \cref{tab:extra_res} in Appendix.}

\vspace{1.5mm}
\noindent \textbf{Performance across Languages.}
The scores over individual target languages of all \textit{translate-test} variants also reveal the presence of ample headroom under English performance in all three tasks. This is due to ``information lost'' owing to imperfect translation. As expected, larger gaps are detected with target languages more dissimilar to English (e.g., compare the scores of \textsc{ja}, \textsc{ko}, and \textsc{zh} versus \textsc{es}, \textsc{fr}, \textsc{de} on PAWS-X), and for lower-resource languages with smaller amounts of parallel data (e.g., \textsc{ht} in XCOPA, \textsc{sw} in XCOPA and XNLI). %

\begin{figure*}
    \centering
    \includegraphics[width=\textwidth]{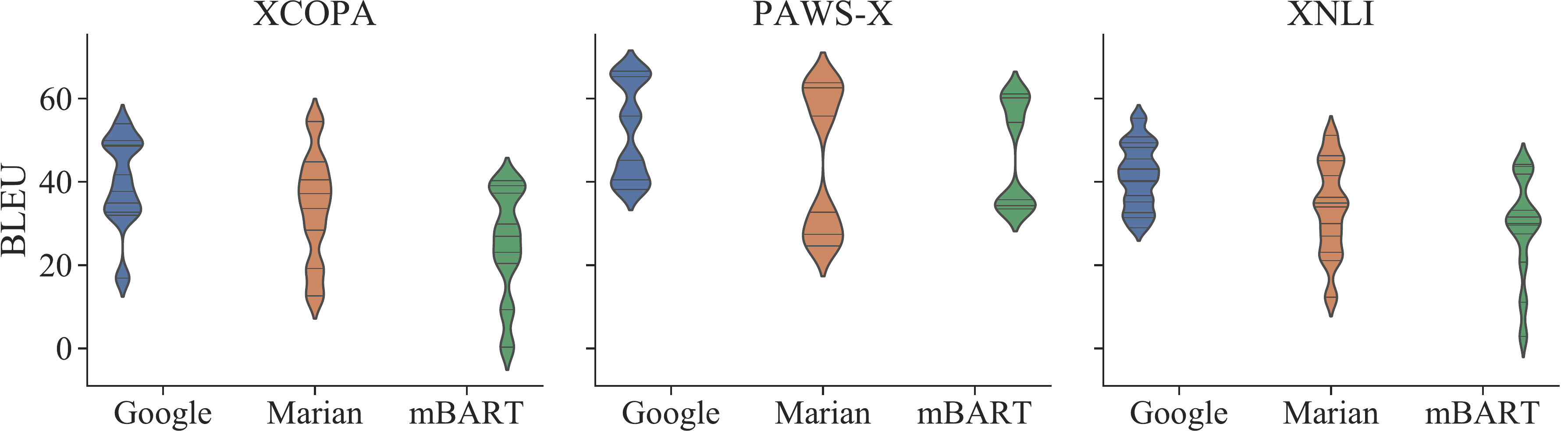}
    \caption{BLEU scores of the 1-best translation for 3 MT models (Google, Marian, and mBART) on the development sets of 3 tasks (XCOPA, PAWS-X, XNLI). Each language is a horizontal line.}
    \label{fig:bleu_dev}
    \vspace{-2mm}
\end{figure*}

\vspace{1.5mm}
\noindent \textbf{Performance across Tasks.} Finally, a cross-task comparison reveals that the largest benefits of the translation-based approaches are observed on the (arguably most complex) XCOPA task. Here, the gains with Google 1-best and Marian 1-best over the ME approach are pronounced in both zero-shot and few-shot setups. Moreover, the use of multiple samples plus MRT yields further performance gains, highest across all tasks. The benefits of the TT approach compared to ME are smaller on XNLI, and non-present on the PAWS-X task, which is the most saturated and least linguistically diverse \citep{ruder2021xtreme}. However, the results across \textit{all} three tasks do confirm the benefits of the proposed latent translation-based approach, which always improves over the base MT system.

\section{In-Depth Analysis}

\begin{figure*}
\centering
      \includegraphics[width=\textwidth]{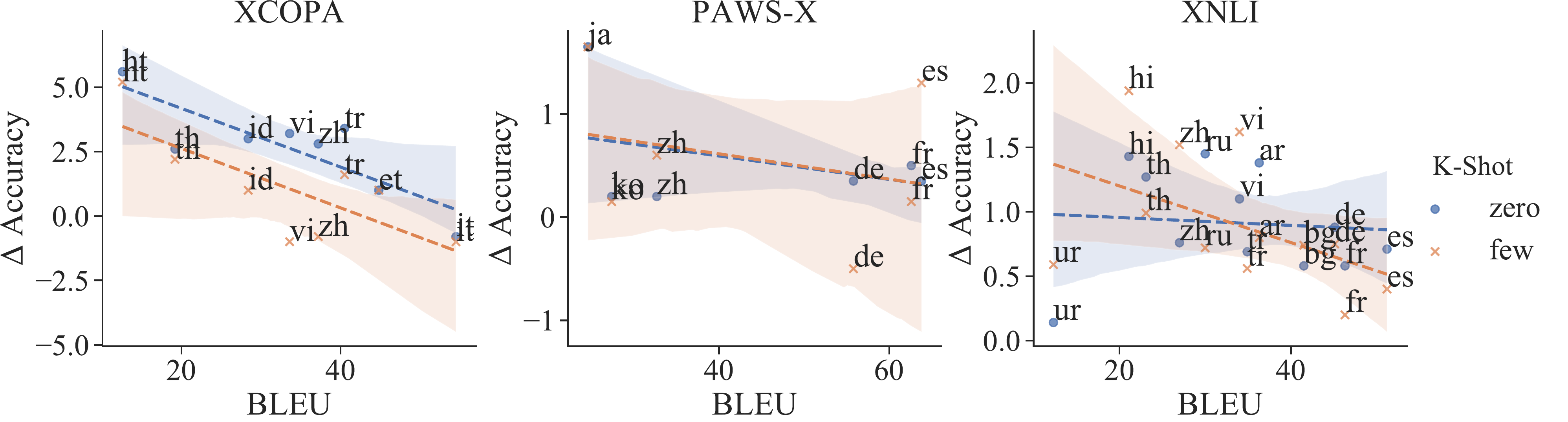}
        \caption{MT quality vs accuracy gains when using 12 rather than 1 translation samples across tasks and languages.}
        \label{fig:mt_gains}
        \vspace{-1.5mm}
\end{figure*}

   \textbf{Number of Samples.} 
   We plot the effect of varying the number of translation samples from 1 to 12 on XCOPA accuracy in \cref{fig:sampling}. As it emerges, the performance metric increases monotonically. The lack of a plateau in the considered interval suggests that larger numbers could ameliorate accuracy even further. Moreover, note that $k$-best deterministic sampling appears superior to probabilistic sampling for all $k$-s.%
   
    \vspace{1.5mm}
    \noindent \textbf{Translation Quality.} In \cref{fig:bleu_dev}, we report the BLEU scores\footnote{Evaluated through \texttt{sacrebleu} \citep{post-2018-call}.} of the 1-best translation of all NMT systems on the development sets of all languages. We source the gold references from the English datasets (COPA, PAWS, and SNLI) from which XCOPA, PAWS-X, and XNLI, respectively, were manually translated. The violin plot reveals large gaps in BLEU based on the distance from English. This is most evident in PAWS-X, with \textsc{ja}, \textsc{ko}, and \textsc{zh} on the bottom and \textsc{de}, \textsc{es}, \textsc{fr} on the top. Moreover, BLEU levels vary by task: while XCOPA has the shortest sentences, it is also the most typologically diverse. This makes the dataset easier to translate in some respects, but harder in others. %
    
    Hence, one might wonder what is the relationship between these 1-best BLEU scores and gains due to multiple samples. %
    Figure~\ref{fig:mt_gains} depicts these two quantities. For XCOPA and XNLI, there is a trend for higher gains with lower translation quality. \textsc{ht} is the clearest example, with only 12.6 BLEU and a $\Delta$ of around 5 points in accuracy.
    However, for PAWS-X we observe almost no linear correlation, which might be caused by overall gains being much smaller. There might also be a minimum translation quality below which multiple samples cannot bring benefits, which would explain the outlier \textsc{ur} in XNLI. On the other extreme, in a few languages with best-quality MT, such as \textsc{it} for XCOPA and \textsc{de} for PAWS-X, multiple samples seem to mislead the classifier in few-shot transfer.
   
   \vspace{1.5mm}
   \noindent \textbf{Translation Ranking.} %
    The fact that lower-scoring translations positively contribute during ensemble prediction is counter-intuitive. Yet, we verify that among the $k$-best translations, higher-ranking ones are not necessarily associated with better classification accuracy, even when their log-probability is significantly greater (see \cref{fig:rank_correctness} in Appendix). Table~\ref{tab:xcopa_example} shows an example for \textsc{tr} in XCOPA, where lower-ranking translations turn the ensemble decision correct when the highest rank translation leads to the wrong classification. In this case, they resolve an incorrect disambiguation of gender. However, in other cases it is hard to interpret how minor lexical changes affect the ensemble prediction.
    
    \begin{table*}[t]
        \centering
        \def\arraystretch{0.93}
        \small
        \begin{tabularx}{\textwidth}{c|lll|c}
        \toprule
          \textbf{NLL} & \textbf{Premise:} Adam partide çok içti. & \textbf{Hyp. 1:} Ertesi gün başı ağrıdı. & \textbf{Hyp. 2:} Ertesi gün burnu aktı. & $\hat{y}$\\
          \midrule
         -1.49 & He drank too much at the party. & She had a headache the next day. & The next day, he had a runny nose. & 1\\
         -2.07 & The guy drank too much at the party. & He had a headache the next day. & The next day, his nose leaked. & 0\\
         -2.26 & The guy drank a lot at the party. & He got a headache the next day. & The next day, his nose ran. & 0 \\
                     \bottomrule
        \end{tabularx}%
        \vspace{-1mm}
        \caption{Marian MT translations from ranks 1-3 for XCOPA \textsc{tr} with model scores (NLL) and classifier predictions ($\hat{y}$). The prompt for this example is: ``What happened as a result?'' and the correct label is 0.}
        \label{tab:xcopa_example}
        \vspace{-1mm}
    \end{table*}

 \vspace{0.2cm}
 \noindent
 These insights offer a provocative question for future work: can similar \textit{application-oriented evaluations} %
 of MT systems reach beyond standard intrinsic evaluation protocols such as BLEU \cite{papineni-etal-2002-bleu} or METEOR \cite{banerjee-lavie-2005-meteor}? In other words, assessing how well the NMT models support cross-lingual transfer might provide additional empirical evidence on their translation abilities. %

\section{Conclusion and Future Work}
We proposed a new method to perform translation-based cross-lingual transfer, by treating the translation of the input text in a target language as a latent random variable. This unifies under a single model both components (a translator and a classifier) of the traditional pipeline, which were previously learned separately and deployed in consecutive steps. As a consequence, in our model, 1) multiple translations can be generated with Monte Carlo sampling to better render the original meaning and 2) the translator can be adapted to the downstream task and correct its errors based on the feedback from the classifier through a variant of Minimum Risk Training. We demonstrate the effectiveness of our method on several benchmarks for natural language understanding, including commonsense causal reasoning, paraphrase identification, and natural language inference. Furthermore, we find that classification performance varies dramatically according to the translation quality of the underlying translator model, whereas its internal ranking of $k$-best translations plays almost no role.
We hope that our findings will provide an incentive to improve language coverage and quality of (especially public) NMT models to support a wide array of multilingual NLP applications.

\bibliography{anthology,emnlp2020}
\bibliographystyle{acl_natbib}

\clearpage
\appendix

 \begin{figure*}[ht]
    \centering
    \includegraphics[width=\textwidth]{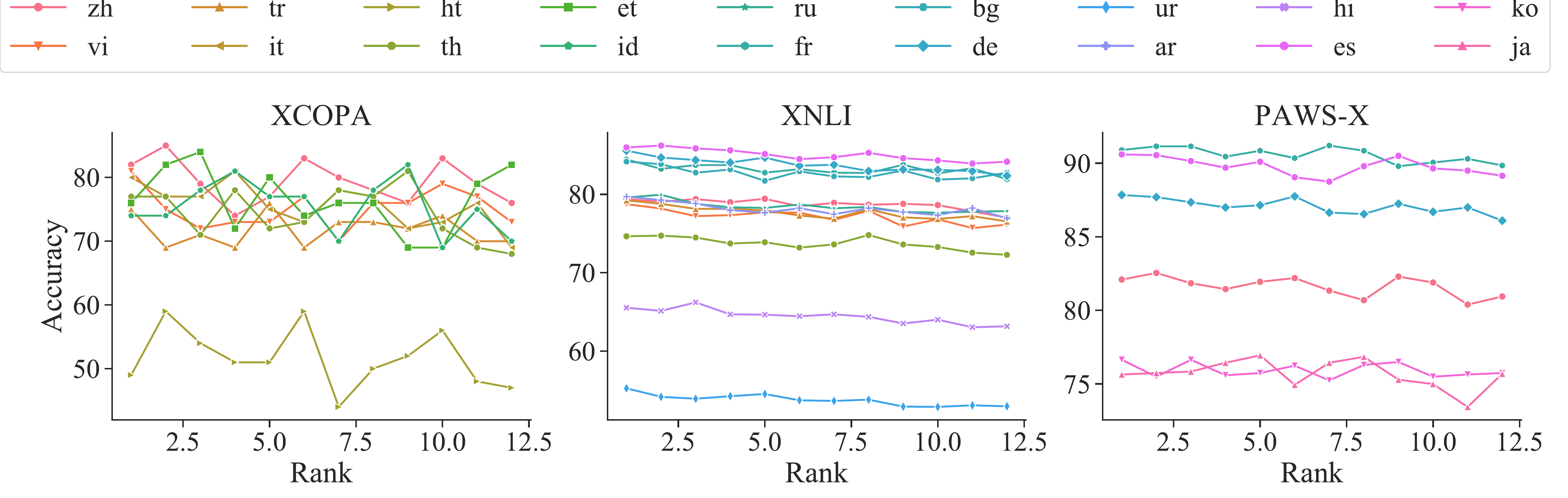}
    \caption{Classification accuracy based on individual translations of different rank in all tasks.}
    \label{fig:rank_correctness}
\end{figure*}

\section{Translation Quality Across Ranks}

Figure~\ref{fig:rank_correctness} shows how in XNLI the share of correct predictions are relatively stable across translations of different ranks within each language. For some languages the top-raking translations are the best candidate (e.g. \textsc{es}, \textsc{ur}), but for others it may be lower-ranked translation (e.g. 3rd best for \textsc{hi} or 8th best for \textsc{th}).

\section{Additional Results}

\begin{table}[h]
    \centering
    \footnotesize
\begin{tabularx}{\columnwidth}{>{\sc}Y|a|YY}
\toprule
& \multicolumn{1}{a|}{\textit{ME}} & \multicolumn{2}{c}{\textit{TT}: RoBERTa +} \\
\hline
    & {\bf mBART} & \textbf{Marian} & \textbf{Marian} \\
& & +Self-train & +Re-ranker \\
    \hline
en  &   77.0 &  89.6 & 89.6   \\
\hline
et  &    56.2 & 86.6 & 70.4  \\
ht  &   $^\star$  & 58.6 & 59.6   \\
id  &  72.8  & 86.6 & 59.4 \\
it  &  61.2 & 85.4 & 64.6       \\
qu  &  $^\star$ & $^\star$ &  $^\star$     \\
sw  &  54.4 & $^\star$ & $^\star$  \\
ta  &  61.4 & $^\star$ & $^\star$  \\
th  & 67.8 & 78.4 & 59.4  \\
tr  &  64.6 & 84.8 & 77.2  \\
vi  &  67.6 & 78.0 & 60.6   \\
zh  &  67.4 & 85.4 & 64.8   \\
\hline
\textbf{\normalfont{avg}} & 63.7 & 79.3 & 64.5 \\
\bottomrule
\end{tabularx}
    \caption{Additional few-shot learning results on XCOPA for alternative multilingual encoders pre-trained on NMT (mBART) and alternative auxiliary objectives (self-training and re-ranking).}
    \label{tab:extra_res}
\end{table}

\section{Related Work}
Minimum Risk Training (MRT) is a technique used for tuning MT models towards given evaluation metrics, and was introduced for statistical MT models~\citep{och-2003-minimum, arun-etal-2010-unified, smith-eisner-2006-minimum} and adapted for NMT~\citep{shen-etal-2016-minimum, edunov-etal-2018-classical, wieting-etal-2019-beyond, wang-sennrich-2020-exposure, saunders-etal-2020-using}. The advantages over classic maximum likelihood training with reference translations are that it mitigates exposure bias and addresses the loss-evaluation mismatch \citep{RanzatoETAL:16, wiseman-rush-2016-sequence, wang-sennrich-2020-exposure} by incorporating the evaluation metric directly into the loss, rewarding high-scoring model outputs and penalizing lower-scoring ones. This metric does not need to be differentiable (e.g. BLEU), and gradients are approximated with Monte-Carlo sampling. MRT has been found effective not only in NMT, but other sequence-to-sequence NLP tasks such as abstractive summarization~\citep{edunov-etal-2018-classical, makino-etal-2019-global}, string transduction~\citep{makarov-clematide-2018-neural}, and referring expression generation~\citep{panagiaris-etal-2020-improving}. While each task comes with its own evaluation metrics, the rewards can also be received from other neural models~\citep{he2016dual, wieting-etal-2019-beyond}, user feedback~\citep{kreutzer-etal-2018-neural}, or a downstream task such as the success of execution of a semantic parse~\citep{misra-etal-2018-policy, jehl-etal-2019-learning} or cross-lingual information retrieval~\citep{sokolov2014}. The latter works are similar to ours in that we leverage downstream task signals to adapt the MT model.

\section{Details for Reproducibility}
We carried out all our experiments on a single 48GB RTX 8000 GPU with Turing architecture. On average, runtime is $\sim3$ hours for fine-tuning on the English training set, $\sim10$ minutes per language for fine-tuning on the target language development set (few-shot learning), and $\sim5$ minutes per language for evaluation.

\end{document}